\documentclass[11pt,a4paper]{article}
\usepackage[hyperref]{emnlp-ijcnlp-2019}
\usepackage{times}
\usepackage{latexsym}

\usepackage{url}
\usepackage{graphicx}
\usepackage{float}
\usepackage{amsmath}
\usepackage{amssymb}
\usepackage{multirow}
\usepackage{booktabs}
\usepackage{url}
\usepackage{array}
\usepackage{algorithm}
\usepackage{algorithmic}
\usepackage{footnote}
\usepackage{enumitem}
\makesavenoteenv{table}
\usepackage{CJKutf8}

\aclfinalcopy 

\title{A Span-Extraction Dataset for Chinese Machine Reading Comprehension}
\author{Yiming Cui$^\dag$$^\ddag$, Ting Liu$^\dag$, Wanxiang Che$^\dag$, \\
\textbf{Li Xiao$^\ddag$, Zhipeng Chen$^\ddag$, Wentao Ma$^\ddag$$^\S$, Shijin Wang$^\ddag$$^\S$, Guoping Hu$^\ddag$} \\
{$^\dag$Research Center for Social Computing and Information Retrieval (SCIR),}\\
{Harbin Institute of Technology, Harbin, China}\\
{$^\ddag$State Key Laboratory of Cognitive Intelligence, iFLYTEK Research, China} \\
{$^\S$iFLYTEK AI Research (Hebei), Langfang, China} \\
{$^\dag$\tt\{ymcui,tliu,car\}@ir.hit.edu.cn}\\
{$^\ddag$$^\S$\tt\{ymcui,lixiao3,zpchen,wtma,sjwang3,gphu\}@iflytek.com}\\  
}

\date{}

\begin{document}
\begin{CJK*}{UTF8}{gbsn}
\maketitle
\begin{abstract}
Machine Reading Comprehension (MRC) has become enormously popular recently and has attracted a lot of attention.
However, the existing reading comprehension datasets are mostly in English.
In this paper, we introduce a Span-Extraction dataset for Chinese machine reading comprehension to add language diversities in this area.
The dataset is composed by near 20,000 real questions annotated on Wikipedia paragraphs by human experts. 
We also annotated a challenge set which contains the questions that need comprehensive understanding and multi-sentence inference throughout the context.
We present several baseline systems as well as anonymous submissions for demonstrating the difficulties in this dataset.
With the release of the dataset, we hosted the Second Evaluation Workshop on Chinese Machine Reading Comprehension (CMRC 2018). 
We hope the release of the dataset could further accelerate the Chinese machine reading comprehension research.\footnote{Resources are available: \url{https://github.com/ymcui/cmrc2018}.}
\end{abstract}


\section{Introduction}
To read and comprehend natural languages is the key to achieve advanced artificial intelligence.
Machine Reading Comprehension (MRC) aims to comprehend the context of given articles and answer the questions based on them. 
Various types of machine reading comprehension datasets have been proposed, such as cloze-style reading comprehension \citep{hermann-etal-2015, hill-etal-2015,cui-etal-2016}, span-extraction reading comprehension \citep{rajpurkar-etal-2016,trischler2016newsqa}, open-domain reading comprehension \citep{nguyen2016ms,he2017dureader}, reading comprehension with multiple-choice \citep{richardson2013mctest,lai2017race}, etc. Along with the development of the reading comprehension dataset, various neural network approaches have been proposed and made a big advancement in this area \citep{kadlec-etal-2016,cui-acl2017-aoa,dhingra-etal-2017,wang-and-jiang-2016,xiong-etal-2016,liu-etal-2017,rnet-2017,mnemonic-2018,slqa-2018,qanet-2018}.

\begin{figure*}[tp]
\centering\scriptsize
        \begin{tabular}{p{0.95\columnwidth} p{0.95\columnwidth}}
        \toprule
	{\bf [Passage]}  & {\bf [Passage]} \\ 
	《黄色脸孔》是柯南·道尔所著的福尔摩斯探案的56个短篇故事之一，收录于《福尔摩斯回忆录》。孟罗先生素来与妻子恩爱，但自从最近邻居新入伙后，孟罗太太则变得很奇怪，曾经凌晨时份外出，又藉丈夫不在家时偷偷走到邻居家中。于是孟罗先生向福尔摩斯求助，福尔摩斯听毕孟罗先生的故事后，认为孟罗太太被来自美国的前夫勒索，所以不敢向孟罗先生说出真相，所以吩咐孟罗先生，如果太太再次走到邻居家时，即时联络他，他会第一时间赶到。孟罗太太又走到邻居家，福尔摩斯陪同孟罗先生冲入，却发现{\bf 邻居家中的人是孟罗太太与前夫生的女儿，因为孟罗太太的前夫是黑人，她怕孟罗先生嫌弃混血儿，所以不敢说出真相。}
	&
	"The Adventure of the Yellow Face", one of the 56 short Sherlock Holmes stories written by Sir Arthur Conan Doyle, is the third tale from The Memoirs of Sherlock Holmes. Mr. Munro has always been loved by his wife, but since the new neighbors recently joined, Mrs. Munro has become very strange. She used to go out in the early hours of the morning and secretly went to her neighbors when her husband was not at home. 
	...
	Mrs. Munro went to the neighbor's house again, and Holmes accompanied Mr. Munro to rush in, only to find that the neighbor's family was the daughter of Mrs. Munro and her ex-husband, {\bf because Mrs. Munro's ex-husband was black, and she was afraid of Mr. Munro hate the mixed-race, so she did not dare to tell the truth.} \\
        \midrule
	{\bf [Question]} & {\bf [Question]} \\
	孟罗太太为什么在邻居新入伙后变得很奇怪？ & 
	Why Mrs. Munro became strange after the new neighbors moved in? \\
	{\bf [Answer 1]} & {\bf [Answer 1]} \\
	邻居家中的人是孟罗太太与前夫生的女儿，因为孟罗太太的前夫是黑人，她怕孟罗先生嫌弃混血儿 & 
	because Mrs. Munro's ex-husband was black, and she was afraid of Mr. Munro hate the mixed-race \\
	{\bf [Answer 2]} & {\bf [Answer 2]} \\
	邻居家中的人是孟罗太太与前夫生的女儿，因为孟罗太太的前夫是黑人，她怕孟罗先生嫌弃混血儿，所以不敢说出真相。 & 
	because Mrs. Munro's ex-husband was black, and she was afraid of Mr. Munro hate the mixed-race\\
	{\bf [Answer 3]} & {\bf [Answer 3]} \\
	邻居家中的人是孟罗太太与前夫生的女儿，因为孟罗太太的前夫是黑人，她怕孟罗先生嫌弃混血儿，所以不敢说出真相。 &
	because Mrs. Munro's ex-husband was black, and she was afraid of Mr. Munro hate the mixed-race, so she did not dare to tell the truth. \\
        \bottomrule
        \end{tabular}
\caption{\label{cmrc2018-example} An example of the proposed CMRC 2018 dataset (challenge set). English translation is also given for comparison.}
\end{figure*}

We also have seen various efforts on the construction of Chinese machine reading comprehension datasets.
In cloze-style reading comprehension, \citet{cui-etal-2016} proposed a Chinese cloze-style reading comprehension dataset: People's Daily \& Children's Fairy Tale.
To add difficulties to the dataset, along with the automatically generated evaluation sets (development and test), they also release a human-annotated evaluation set. 
Later, \citet{cmrc2017-dataset} propose another dataset, which is gathered from children's reading material. To add more diversity and for further investigation on transfer learning, they also provide another evaluation dataset, which is also annotated by human experts, but the query is more natural than the cloze type. The dataset was used in the first evaluation workshop on Chinese machine reading comprehension (CMRC 2017).
In open-domain reading comprehension, \citet{he2017dureader} propose a large-scale open-domain Chinese machine reading comprehension dataset (DuReader), which contains 200k queries annotated from the user query logs on the search engine.
\citet{shao2018drcd} proposed a reading comprehension dataset in Traditional Chinese.

Though we have seen that the current machine learning approaches have surpassed the human performance on the SQuAD dataset \cite{rajpurkar-etal-2016}, we wonder if these state-of-the-art models could also give a similar performance on the dataset of different languages. To further accelerate the development of the machine reading comprehension research, we propose a span-extraction dataset for Chinese machine reading comprehension.
Figure \ref{cmrc2018-example} shows an example of the proposed dataset.
The main contributions of our work can be concluded as follows.
\begin{itemize}[leftmargin=*]
	\item We propose a Chinese span-extraction reading comprehension dataset which contains near 20,000 human-annotated questions, to add linguistic diversity in reading comprehension field.
	\item To thoroughly test the ability of the MRC systems, besides the development and test set, we also make a challenge set which contains carefully annotated questions that require various clues in the passage. The BERT-based approaches could only give under 50\% F1-score on this set, indicating its difficulty.
	\item The proposed Chinese RC data could also be a resource for cross-lingual research purpose when studied along with SQuAD and other similar datasets.
\end{itemize}

\section{The Proposed Dataset}\label{dataset}
\subsection{Task Definition}
Generally, the reading comprehension task can be described as a triple $\langle \mathcal P, \mathcal Q, \mathcal A \rangle$, where $\mathcal P$ represents {\bf P}assage, $\mathcal Q$ represents {\bf Q}uestion and the $\mathcal A$ represents {\bf A}nswer. 
Specifically, for span-extraction reading comprehension task, the question is annotated by the human, which is much more natural than the cloze-style MRC datasets \citep{hill-etal-2015,cui-etal-2016}. The answer $\mathcal A$ should be a span which is directly extracted from the passage $\mathcal P$.
According to most of the works on SQuAD, the task can be simplified by predicting the start and end pointer in the passage \citep{wang-and-jiang-2016}.

\subsection{Data Pre-Processing}
We downloaded Chinese portion of Wikipedia webpage dump\footnote{\url{https://dumps.wikimedia.org/zhwiki/latest/}} on Jan 22, 2018 and used open-source toolkit {\em Wikipedia Extractor}\footnote{\url{http://medialab.di.unipi.it/wiki/Wikipedia_Extractor}} for pre-processing the raw files into plain text. 
We also convert the Traditional Chinese characters into Simplified Chinese for normalization purpose using {\em opencc\footnote{\url{https://github.com/BYVoid/OpenCC}}} toolkit.

\subsection{Human Annotation}
The questions in the proposed dataset are completely annotated by human experts, which is different from previous works that rely on the automatic data generation \citep{hermann-etal-2015,hill-etal-2015,cui-etal-2016}. Before annotating, the document is divided into several passages, and each passage is limited to have no more than 500 Chinese words, where the word is counted by using LTP \citep{che2010ltp}. Then, the annotator was instructed to first evaluate the appropriateness of the passages, because some of the passages are extremely difficult for the public to understand. Following rules are applied when discarding the passages.
\begin{itemize}[leftmargin=*]
	\item Contain over 30\% non-Chinese characters.
	\item Contain many professional words that hard to understand. 
	\item Contains many special characters and symbols.
	\item The paragraph is written in classical Chinese, which is substantially different from the Chinese language nowadays.
\end{itemize}

After identifying the passage is appropriate for annotation, the annotator will read the passage and ask the questions based on it and annotated a primary answer.
During the question annotation, the following rules are applied.
\begin{itemize}[leftmargin=*]
	\item No more than five questions for each passage.
	\item The answer MUST be a span in the passage to meet the task definition.
	\item Encourage the question diversity, such as who/when/where/why/how, etc.
	\item Avoid directly using the description in the passage. Use paraphrase or syntax transformation to add difficulties for answering.
	\item Long answers (say over 30 characters) will be discarded.
\end{itemize}

For the evaluation sets, i.e., development, test, challenge, there are three answers available for better evaluation. 
Besides the primary answer that was annotated by the question proposer, we also invite two additional annotators to write the second and third answers for the question. During this phase, the annotators could not see the primary answer to ensure the answer was not copied from others and encourage the diversities in the answer.

\subsection{Challenge Set}
In order to examine how well can reading comprehension models deal with the questions that need comprehensive reasoning over various clues in the context, we additionally annotated a small challenge set for this purpose while keeping the span-extraction style. 
The annotation was also done by three annotators in a similar way that for development and test set. 
Figure \ref{cmrc2018-example} shows an example in the challenge set.
The question should meet the following standards to be qualified into this set.
\begin{itemize}[leftmargin=*]
	\item The answer cannot be only inferred by a single sentence in the passage if the answer is a single word or short phrase. We encourage the annotator to ask the questions that need comprehensive reasoning in the passage to increase the difficulties.
	\item If the answer belongs to a type of named entity, or specific genre (such as date, color, etc.), it can not be the only one in the context, or the machine could easily pick it out according to its type. For example, if there is only one person name appears in the context, then it cannot be used for annotating questions. There should be at least two person names that could mislead the machine for answering.
\end{itemize}

\subsection{Statistics}
\begin{figure}[tp]
  \centering
  \includegraphics[width=0.48\textwidth]{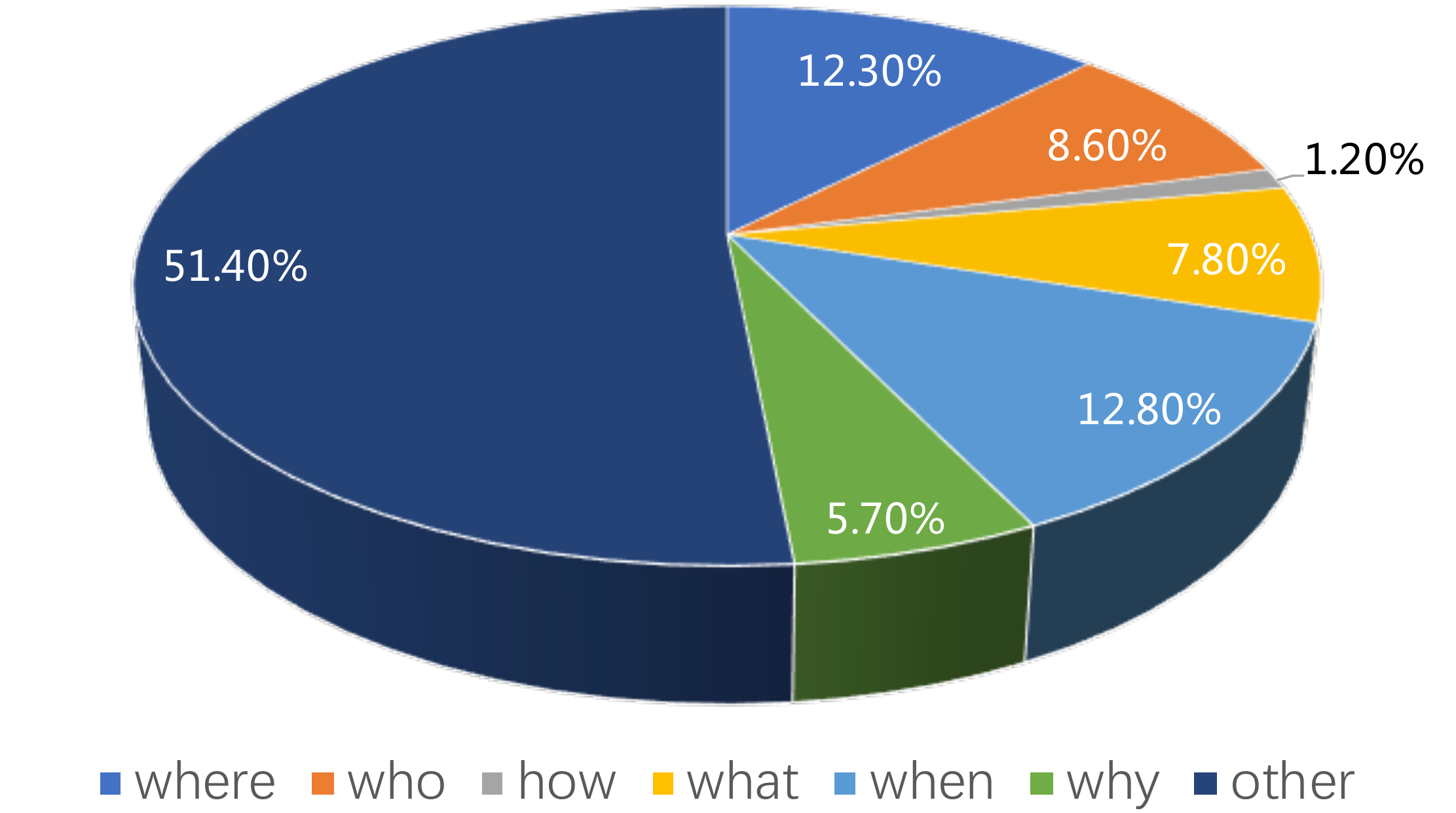}
  \caption{\label{query-type} Question types of the development set.}
\end{figure}

The general statistics of the pre-processed data are given in Table \ref{data-stats}. The question type distribution of the development set is given in Figure \ref{query-type}.

\begin{table}[ht]
\small
\begin{center}
\begin{tabular}{l rrr | r}
\toprule
 & \bf Train & \bf Dev & \bf Test & \bf Challenge \\
\midrule
Question \# & 10,321 & 3,351 & 4,895 & 504 \\
Answer \# per query & 1 & {\bf 3} & {\bf 3} & {\bf 3} \\
Max passage tokens & 962 & 961 & 980 & 916 \\
Max question tokens & 89 & 56 & 50  & 47 \\
Max answer tokens & 100 & 85 & 92  &77 \\
Avg passage tokens & 452 & 469 & 472  & 464 \\
Avg question tokens & 15 & 15 & 15 & 18 \\
Avg answer tokens & 17 & 9 & 9  & 19 \\
\bottomrule
\end{tabular}
\end{center}
\caption{\label{data-stats} Statistics of the CMRC 2018 dataset.}
\end{table}

        \begin{table*}[htp]
        \begin{center}
        \small
        \begin{tabular}{l  cccc | cc}
        \toprule
        & \multicolumn{2}{c}{\bf Development} & \multicolumn{2}{c}{\bf Test} & \multicolumn{2}{c}{\bf Challenge}\\
        & EM & F1 & EM & F1 & EM & F1 \\
        \midrule
        {\em Estimated Human Performance} & \em 91.083 & \em 97.348 & \em 92.400 & \em 97.914 & \em 90.382 & \em 95.248 \\
        \midrule 
        Z-Reader (single model) & \bf 79.776 & \bf 92.696 & \bf 74.178 & \bf 88.145 & 13.889 & \bf 37.422 \\
        MCA-Reader (ensemble) & 66.698 & 85.538 & 71.175 & 88.090 & 15.476 & 37.104 \\
        RCEN (ensemble) & 76.328 & 91.370 & 68.662 & 85.753 & 15.278 & 34.479 \\
        MCA-Reader (single model) & 63.902 & 82.618 & 68.335 & 85.707 & 13.690 & 33.964 \\
        OmegaOne (ensemble) & 66.977 & 84.955 & 66.272 & 82.788 & 12.103 & 30.859 \\ 
        RCEN (single model) & 73.253 & 89.750 & 64.576 & 83.136 & 10.516 & 30.994 \\
        GM-Reader (ensemble) & 58.931 & 80.069 & 64.045 & 83.046 & \bf 15.675 & 37.315 \\ 
        OmegaOne (single model) & 64.430 & 82.699 & 64.188 & 81.539 & 10.119 & 29.716 \\ 
        GM-Reader (single model) & 56.322 & 77.412 & 60.470 & 80.035 & 13.690 & 33.990 \\ 
        R-NET (single model) & 45.418 & 69.825 & 50.112 & 73.353 & 9.921 & 29.324 \\ 
        SXU-Reader (ensemble) & 40.292 & 66.451 & 46.210 & 70.482 & N/A & N/A \\ 
        SXU-Reader (single model) & 37.310 & 66.121 & 44.270 & 70.673 & 6.548 & 28.116 \\         
        T-Reader (single model) & 39.422 & 62.414 & 44.883 & 66.859 & 7.341 & 22.317 \\        
        \midrule
        \bf BERT-base (Chinese) & 63.6 & 83.9 & 67.8 & 86.0 & 18.4 & 42.1 \\
        \bf BERT-base (Multi-lingual) & 64.1 & 84.4 & 68.6 & 86.8 & 18.6  & 43.8 \\
        \bottomrule
        \end{tabular}
        \end{center}
        \caption{\label{result-baseline} Baseline results and CMRC 2018 participants' results. Note that, some of the submissions are using development set for training as well.}
        \end{table*}

\section{Evaluation Metrics}
In this paper, we adopt two evaluation metrics following \citet{rajpurkar-etal-2016}.
However, as the Chinese language is quite different from English, we adapt the original metrics in the following ways. 
Note that, the common punctuations, white spaces are ignored for normalization.

\subsection{Exact Match}
Measure the exact match between the prediction and ground truths that is 1 for the exact match. Otherwise, the score is 0.
This is the same as the one proposed by \citet{rajpurkar-etal-2016}.

\subsection{F1-Score}
Measure the character-level fuzzy match between the prediction and ground truths.
Instead of treating the predictions and ground truths as bag-of-words, we calculate the length of the longest common sequence (LCS) between them and compute the F1-score accordingly. 
We take the maximum F1 over all of the ground truth answers for a given question. Note that, non-Chinese words will not be segmented into characters.

\subsection{Estimated Human Performance}
We also report the estimated human performance in order to measure the difficulty of the proposed dataset.
As we have illustrated in the previous section, there are three answers for each question in development, test, and challenge set. Unlike \citet{rajpurkar-etal-2016}, we use a cross-validation method to calculate the performance. 
We regard the first answer as human prediction and treat the rest of the answers as ground truths. 
In this way, we can get three human prediction performance by iteratively regarding the first, second, and third answer as the human prediction.
Finally, we calculate the average of three results as the final estimated human performance on this dataset.

\section{Experimental Results}
\subsection{Baseline System}
Following \citet{devlin2018bert}, we adopt BERT for our baseline system.
Specifically, we slightly modify the {\tt run\_squad.py} script\footnote{\url{https://github.com/google-research/bert/blob/master/run_squad.py}} for adjusting our dataset, while keeping the most of the original implementation.
For the baseline system, we used an initial learning rate of 3e-5 with a batch size of 32 and trained for two epochs.
The maximum lengths of document and query are set to 512 and 64.

\subsection{Results}
The results are shown in Table \ref{result-baseline}.
Besides the baseline systems, we also include the participants' results of CMRC 2018 evaluation.
We release the training and development set to the public and accepted submissions from participants to evaluate their models on the hidden test and challenge set to preserve the integrity of the evaluation process following \citet{rajpurkar-etal-2016}.
As we can see that most of the participants could obtain over 80 in the test F1.
While compared to F1 metric, the EM metric is substantially lower compared to the SQuAD dataset (usually within 10 points). 
This suggests that how to determine the exact span boundary in Chinese machine reading comprehension plays a key role to improve the system performance.

Not surprisingly, as shown in the last column of Table \ref{result-baseline}, though the top-ranked systems obtain decent scores on the development and test set, they are failed to give satisfactory results on the challenge set. 
However, as we can see that the estimated human performance on the development, test, and challenge set are relatively similar, where the challenge set gives slightly lower scores. 
We also observed that though Z-Reader obtains best scores on the test set, it failed to give consistent performances on the EM metric of the challenge set. 
This suggests that the current reading comprehension models are relatively not capable of handling difficult questions that need comprehensive reasoning among several clues in the passage.

BERT-based approaches show competitive performance against participants submissions.\footnote{As CMRC 2018 workshop was held before the publication of BERT, systems of the participants are not based on BERT.} 
Though traditional models have higher scores in the test set, when it comes to the challenge set, BERT-based baselines are consistently higher, demonstrating that rich representation provided by BERT is beneficial for solving harder questions and generalize well among both easy and hard questions.

\section{Conclusion}\label{conclusion}
In this work, we propose a span-extraction dataset for Chinese machine reading comprehension. 
The dataset is annotated by human experts with near 20,000 questions as well as a challenging set which is composed of the questions that need reasoning over multiple clues. 
The evaluation results show that the machine could give excellent scores on the development and test set with only near 10 points below the estimated human performance in F1-score. However, when it comes to the challenge set, the scores are declining drastically while the human performance remains almost the same with the non-challenge set, indicating that there are still potential challenges in designing more sophisticated models to improve the performance.
We hope the release of this dataset could bring language diversity in machine reading comprehension task, and accelerate further investigation on solving the questions that need comprehensive reasoning over multiple clues.

\section*{Open Challenge}
We would like to invite more researchers doing experiments on our CMRC 2018 dataset and evaluate on the hidden test and challenge set to further test the generalization of the models. 
You can follow the instructions on our CodaLab Worksheet to submit your model via {\url{https://bit.ly/2ZdS8Ct}}

\section*{Acknowledgments}
We would like to thank the anonymous reviewers for their thorough reviewing and providing thoughtful comments to improve our paper. 
We would like to thank our resource team for annotating and verifying evaluation data. 
Also, we thank the Seventeenth China National Conference on Computational Linguistics (CCL 2018)\footnote{\url{http://www.cips-cl.org/static/CCL2018/index.html}} and Changsha University of Science and Technology for providing venue for evaluation workshop. 
This work was supported by the National Natural Science Foundation of China (NSFC) via grant 61976072, 61632011 and 61772153.

\bibliography{emnlp-ijcnlp-2019}

\begin{thebibliography}{23}
\expandafter\ifx\csname natexlab\endcsname\relax\def\natexlab#1{#1}\fi

\bibitem[{Che et~al.(2010)Che, Li, and Liu}]{che2010ltp}
Wanxiang Che, Zhenghua Li, and Ting Liu. 2010.
\newblock Ltp: A chinese language technology platform.
\newblock In \emph{Proceedings of the 23rd International Conference on
  Computational Linguistics: Demonstrations}, pages 13--16. Association for
  Computational Linguistics.

\bibitem[{Cui et~al.(2017)Cui, Chen, Wei, Wang, Liu, and Hu}]{cui-acl2017-aoa}
Yiming Cui, Zhipeng Chen, Si~Wei, Shijin Wang, Ting Liu, and Guoping Hu. 2017.
\newblock \href {https://doi.org/10.18653/v1/P17-1055}
  {Attention-over-attention neural networks for reading comprehension}.
\newblock In \emph{Proceedings of the 55th Annual Meeting of the Association
  for Computational Linguistics (Volume 1: Long Papers)}, pages 593--602.
  Association for Computational Linguistics.

\bibitem[{Cui et~al.(2018)Cui, Liu, Chen, Ma, Wang, and Hu}]{cmrc2017-dataset}
Yiming Cui, Ting Liu, Zhipeng Chen, Wentao Ma, Shijin Wang, and Guoping Hu.
  2018.
\newblock Dataset for the first evaluation on chinese machine reading
  comprehension.
\newblock In \emph{Proceedings of the Eleventh International Conference on
  Language Resources and Evaluation (LREC 2018)}. European Language Resources
  Association (ELRA).

\bibitem[{Cui et~al.(2016)Cui, Liu, Chen, Wang, and Hu}]{cui-etal-2016}
Yiming Cui, Ting Liu, Zhipeng Chen, Shijin Wang, and Guoping Hu. 2016.
\newblock \href {http://aclweb.org/anthology/C16-1167} {Consensus
  attention-based neural networks for chinese reading comprehension}.
\newblock In \emph{Proceedings of COLING 2016, the 26th International
  Conference on Computational Linguistics: Technical Papers}, pages 1777--1786.
  The COLING 2016 Organizing Committee.

\bibitem[{Devlin et~al.(2019)Devlin, Chang, Lee, and
  Toutanova}]{devlin2018bert}
Jacob Devlin, Ming-Wei Chang, Kenton Lee, and Kristina Toutanova. 2019.
\newblock \href {https://doi.org/10.18653/v1/N19-1423} {{BERT}: Pre-training of
  deep bidirectional transformers for language understanding}.
\newblock In \emph{Proceedings of the 2019 Conference of the North {A}merican
  Chapter of the Association for Computational Linguistics: Human Language
  Technologies, Volume 1 (Long and Short Papers)}, pages 4171--4186,
  Minneapolis, Minnesota. Association for Computational Linguistics.

\bibitem[{Dhingra et~al.(2017)Dhingra, Liu, Yang, Cohen, and
  Salakhutdinov}]{dhingra-etal-2017}
Bhuwan Dhingra, Hanxiao Liu, Zhilin Yang, William Cohen, and Ruslan
  Salakhutdinov. 2017.
\newblock \href {https://doi.org/10.18653/v1/P17-1168} {Gated-attention readers
  for text comprehension}.
\newblock In \emph{Proceedings of the 55th Annual Meeting of the Association
  for Computational Linguistics (Volume 1: Long Papers)}, pages 1832--1846.
  Association for Computational Linguistics.

\bibitem[{He et~al.(2017)He, Liu, Lyu, Zhao, Xiao, Liu, Wang, Wu, She, Liu
  et~al.}]{he2017dureader}
Wei He, Kai Liu, Yajuan Lyu, Shiqi Zhao, Xinyan Xiao, Yuan Liu, Yizhong Wang,
  Hua Wu, Qiaoqiao She, Xuan Liu, et~al. 2017.
\newblock Dureader: a chinese machine reading comprehension dataset from
  real-world applications.
\newblock \emph{arXiv preprint arXiv:1711.05073}.

\bibitem[{Hermann et~al.(2015)Hermann, Kocisky, Grefenstette, Espeholt, Kay,
  Suleyman, and Blunsom}]{hermann-etal-2015}
Karl~Moritz Hermann, Tomas Kocisky, Edward Grefenstette, Lasse Espeholt, Will
  Kay, Mustafa Suleyman, and Phil Blunsom. 2015.
\newblock Teaching machines to read and comprehend.
\newblock In \emph{Advances in Neural Information Processing Systems}, pages
  1684--1692.

\bibitem[{Hill et~al.(2015)Hill, Bordes, Chopra, and Weston}]{hill-etal-2015}
Felix Hill, Antoine Bordes, Sumit Chopra, and Jason Weston. 2015.
\newblock The goldilocks principle: Reading children's books with explicit
  memory representations.
\newblock \emph{arXiv preprint arXiv:1511.02301}.

\bibitem[{Hu et~al.(2018)Hu, Peng, Huang, Qiu, Wei, and Zhou}]{mnemonic-2018}
Minghao Hu, Yuxing Peng, Zhen Huang, Xipeng Qiu, Furu Wei, and Ming Zhou. 2018.
\newblock \href {https://doi.org/10.24963/ijcai.2018/570} {Reinforced mnemonic
  reader for machine reading comprehension}.
\newblock In \emph{Proceedings of the Twenty-Seventh International Joint
  Conference on Artificial Intelligence, {IJCAI-18}}, pages 4099--4106.
  International Joint Conferences on Artificial Intelligence Organization.

\bibitem[{Kadlec et~al.(2016)Kadlec, Schmid, Bajgar, and
  Kleindienst}]{kadlec-etal-2016}
Rudolf Kadlec, Martin Schmid, Ond{\v{r}}ej Bajgar, and Jan Kleindienst. 2016.
\newblock \href {https://doi.org/10.18653/v1/P16-1086} {Text understanding with
  the attention sum reader network}.
\newblock In \emph{Proceedings of the 54th Annual Meeting of the Association
  for Computational Linguistics (Volume 1: Long Papers)}, pages 908--918.
  Association for Computational Linguistics.

\bibitem[{Lai et~al.(2017)Lai, Xie, Liu, Yang, and Hovy}]{lai2017race}
Guokun Lai, Qizhe Xie, Hanxiao Liu, Yiming Yang, and Eduard Hovy. 2017.
\newblock \href {http://aclweb.org/anthology/D17-1082} {Race: Large-scale
  reading comprehension dataset from examinations}.
\newblock In \emph{Proceedings of the 2017 Conference on Empirical Methods in
  Natural Language Processing}, pages 785--794. Association for Computational
  Linguistics.

\bibitem[{Liu et~al.(2017)Liu, Cui, Yin, Zhang, Wang, and Hu}]{liu-etal-2017}
Ting Liu, Yiming Cui, Qingyu Yin, Wei-Nan Zhang, Shijin Wang, and Guoping Hu.
  2017.
\newblock \href {https://doi.org/10.18653/v1/P17-1010} {Generating and
  exploiting large-scale pseudo training data for zero pronoun resolution}.
\newblock In \emph{Proceedings of the 55th Annual Meeting of the Association
  for Computational Linguistics (Volume 1: Long Papers)}, pages 102--111.
  Association for Computational Linguistics.

\bibitem[{Nguyen et~al.(2016)Nguyen, Rosenberg, Song, Gao, Tiwary, Majumder,
  and Deng}]{nguyen2016ms}
Tri Nguyen, Mir Rosenberg, Xia Song, Jianfeng Gao, Saurabh Tiwary, Rangan
  Majumder, and Li~Deng. 2016.
\newblock Ms marco: A human generated machine reading comprehension dataset.
\newblock \emph{arXiv preprint arXiv:1611.09268}.

\bibitem[{Rajpurkar et~al.(2016)Rajpurkar, Zhang, Lopyrev, and
  Liang}]{rajpurkar-etal-2016}
Pranav Rajpurkar, Jian Zhang, Konstantin Lopyrev, and Percy Liang. 2016.
\newblock \href {https://doi.org/10.18653/v1/D16-1264} {Squad: 100,000+
  questions for machine comprehension of text}.
\newblock In \emph{Proceedings of the 2016 Conference on Empirical Methods in
  Natural Language Processing}, pages 2383--2392. Association for Computational
  Linguistics.

\bibitem[{Richardson et~al.(2013)Richardson, Burges, and
  Renshaw}]{richardson2013mctest}
Matthew Richardson, Christopher~JC Burges, and Erin Renshaw. 2013.
\newblock Mctest: A challenge dataset for the open-domain machine comprehension
  of text.
\newblock In \emph{Proceedings of the 2013 Conference on Empirical Methods in
  Natural Language Processing}, pages 193--203.

\bibitem[{Shao et~al.(2018)Shao, Liu, Lai, Tseng, and Tsai}]{shao2018drcd}
Chih~Chieh Shao, Trois Liu, Yuting Lai, Yiying Tseng, and Sam Tsai. 2018.
\newblock Drcd: a chinese machine reading comprehension dataset.
\newblock \emph{arXiv preprint arXiv:1806.00920}.

\bibitem[{Trischler et~al.(2016)Trischler, Wang, Yuan, Harris, Sordoni,
  Bachman, and Suleman}]{trischler2016newsqa}
Adam Trischler, Tong Wang, Xingdi Yuan, Justin Harris, Alessandro Sordoni,
  Philip Bachman, and Kaheer Suleman. 2016.
\newblock Newsqa: A machine comprehension dataset.
\newblock \emph{arXiv preprint arXiv:1611.09830}.

\bibitem[{Wang and Jiang(2016)}]{wang-and-jiang-2016}
Shuohang Wang and Jing Jiang. 2016.
\newblock Machine comprehension using match-lstm and answer pointer.
\newblock \emph{arXiv preprint arXiv:1608.07905}.

\bibitem[{Wang et~al.(2018)Wang, Yan, and Wu}]{slqa-2018}
Wei Wang, Ming Yan, and Chen Wu. 2018.
\newblock \href {http://aclweb.org/anthology/P18-1158} {Multi-granularity
  hierarchical attention fusion networks for reading comprehension and question
  answering}.
\newblock In \emph{Proceedings of the 56th Annual Meeting of the Association
  for Computational Linguistics (Volume 1: Long Papers)}, pages 1705--1714.
  Association for Computational Linguistics.

\bibitem[{Wang et~al.(2017)Wang, Yang, Wei, Chang, and Zhou}]{rnet-2017}
Wenhui Wang, Nan Yang, Furu Wei, Baobao Chang, and Ming Zhou. 2017.
\newblock \href {https://doi.org/10.18653/v1/P17-1018} {Gated self-matching
  networks for reading comprehension and question answering}.
\newblock In \emph{Proceedings of the 55th Annual Meeting of the Association
  for Computational Linguistics (Volume 1: Long Papers)}, pages 189--198.
  Association for Computational Linguistics.

\bibitem[{Xiong et~al.(2016)Xiong, Zhong, and Socher}]{xiong-etal-2016}
Caiming Xiong, Victor Zhong, and Richard Socher. 2016.
\newblock Dynamic coattention networks for question answering.
\newblock \emph{arXiv preprint arXiv:1611.01604}.

\bibitem[{Yu et~al.(2018)Yu, Dohan, Luong, Zhao, Chen, Norouzi, and
  Le}]{qanet-2018}
Adams~Wei Yu, David Dohan, Minh-Thang Luong, Rui Zhao, Kai Chen, Mohammad
  Norouzi, and Quoc~V Le. 2018.
\newblock Qanet: Combining local convolution with global self-attention for
  reading comprehension.
\newblock \emph{arXiv preprint arXiv:1804.09541}.

\end{thebibliography}
\bibliographystyle{acl_natbib}

\end{CJK*}
\end{document}